\documentclass{article}

\usepackage{PRIMEarxiv}

\usepackage[utf8]{inputenc}    
\usepackage[T1]{fontenc}       
\usepackage{hyperref}          
\usepackage{url}               
\usepackage{booktabs}          
\usepackage{amsfonts}          
\usepackage{nicefrac}          
\usepackage{microtype}         
\usepackage{graphicx}          
\usepackage{tabularx}          
\usepackage{fancyhdr}          
\usepackage{lipsum}            

\graphicspath{{media/}}     

\title{MCUBench: A Benchmark of \\Tiny Object Detectors on MCUs}

\author{
  {\bf Sudhakar Sah, Darshan C. Ganji, Matteo Grimaldi,} \\ 
  {\bf Ravish Kumar, Alexander Hoffman, Honnesh Rohmetra, Ehsan Saboori} \\ \\
  Deeplite, Inc.\\
  Toronto, Canada \\
  \texttt{sudhakar@deeplite.ai}
}

\begin{document}
\maketitle

\begin{abstract}
We introduce \textit{MCUBench}, a benchmark featuring over 100 YOLO-based object detection models evaluated on the VOC dataset across seven different MCUs. This benchmark provides detailed data on average precision, latency, RAM, and Flash usage for various input resolutions and YOLO-based one-stage detectors. By conducting a controlled comparison with a fixed training pipeline, we collect comprehensive performance metrics. Our Pareto-optimal analysis shows that integrating modern detection heads and training techniques allows various YOLO architectures, including legacy models like YOLOv3, to achieve a highly efficient tradeoff between mean Average Precision (mAP) and latency. MCUBench serves as a valuable tool for benchmarking the MCU performance of contemporary object detectors and aids in model selection based on specific constraints. Code and data are available at \href{https://github.com/Deeplite/deeplite-torch-zoo}{github.com/Deeplite/deeplite-torch-zoo}.
\end{abstract}


\section{Introduction}
\label{sec:intro}

In the realm of computer vision, object detection is a crucial task involving the recognition and localization of objects within an image. This technology is essential for various applications, including self-driving cars, security systems, robotics, and augmented reality~\cite{yolo_applications}. Recent advancements in artificial intelligence (AI) have significantly propelled these applications forward, primarily due to the increased processing power and memory capacity provided by specialized hardware like GPUs (Graphical Processing Units) for training and NPUs (Neural Processing Units) for inference. However, many computer vision applications demand deployment at the edge, where considerations such as privacy, low latency, and power efficiency are paramount. This creates a challenge in deploying deep learning-based object detection models on tiny hardware platforms like Microcontroller Units (MCUs), which have limited computational resources and memory~\cite{lane2017squeezing}. These constraints necessitate the development of efficient object detection algorithms tailored for low-footprint devices including MCUs, are often preferred for their affordability and widespread availability, despite their constrained hardware capabilities. Consequently, there is a pressing need for innovative object detection models that can deliver optimal performance on these tiny, general-purpose devices, which are often available for just a few dollars.

YOLO (You Only Look Once) models have dominated the field of real-time object detection since their introduction by Joseph Redmon et al. in $2016$~\cite{redmon2016you}. Unlike most previous methods that employed a two-stage approach, YOLO utilizes a single-stage network \cite{ssd} to simultaneously predict bounding boxes and class probabilities directly from full images. Combining these tasks in one pass significantly enhances speed and efficiency. The evolution from YOLOv$1$~\cite{redmon2016you} to the latest YOLOv$8$~\cite{ultralytics} has brought about numerous improvements. New YOLO versions have introduced new efficient backbone networks for feature extraction, refined neck structures for better feature fusion, and detection heads with options for anchor-based~\cite{redmon2016you, redmon2016_v2, redmon2018yolov3, bochkovskiy2020yolov4, Jocher_YOLOv5_by_Ultralytics_2020, li2023yolov6, wang2023yolov7} anchor-free~\cite{ge2021yolox}, and mixed approaches~\cite{ultralytics}. The integration of innovative loss functions~\cite{dfl} and sophisticated training techniques, such as advanced data augmentation~\cite{dataaug} and meticulous hyperparameter optimization, have further boosted performance.

Despite these advancements, deploying YOLO models on resource-constrained devices like Microcontroller Units (MCUs) remains challenging. Most of the MCUs still use Single Shot Detector (SSD) \cite{liu2016ssd} due to the ease of deployment but SSD models are inferior in terms of performance. Even the smallest variants, such as YOLOv$5$n, which is $7.6$ MB in FP$32$ and about $2$ MB when quantized to $8$-bit, are still too large for most MCUs, which typically have less than $2$ MB of Flash and $1$ MB of RAM~\cite{stm32_f769NI}. These strict hardware constraints motivate the development of ultra-compact YOLO variants.

This paper aims to address these challenges by providing a benchmark of more than hundreds of YOLO-based object detector model architectures generated under controlled conditions (e.g., the same training loop for all models) to demonstrate the impact of the backbone and neck structure of YOLO-based models on MCUs. First we train the models on the Pascal VOC dataset, then we measure inference performance on four different Nucleo boards from STMicroelectronics. Finally, we identify the Pareto optimal set of models according to their VOC average precision and on-device latency, released as the benchmark called {\em MCUBench}.

We summarize our contributions as follows:
\begin{itemize}
    \item We introduce {\em MCUBench}, a comprehensive benchmark of over $100$ tiny YOLO-based object detection models specifically designed for MCU-grade hardware. These models are trained on the VOC dataset, and selected through Pareto analysis across four different MCU platforms.
    \item We demonstrate that using modern detection heads and advanced training pipelines on backbones and necks from different YOLO versions reveals unique performance trends on MCUs compared to more powerful hardware like GPUs. Our benchmark highlights that even older models (e.g., YOLOv$3$) equipped with new detection heads can outperform newer models on MCUs.
    \item We provide the trained weights of {\em MCUBench} models, enabling application developers to select and fine-tune models based on their specific average precision, latency, RAM, and Flash trade-offs, without the need to train multiple models independently.
\end{itemize}

\section{Related Work}
\label{sec:related}

There is a growing interest in benchmarking deep learning models on MCUs, as they offer an affordable solution for compact, low-power use cases. However, there are still many open challenges and opportunities for further research, such as improving the model compression and quantization techniques, exploring the trade-offs between accuracy, speed, and energy, and developing more realistic and diverse benchmark models and datasets. 

MLPerf\cite{mlperf}, a community-led benchmarking initiative, offers a benchmarking suite for ML inference. However, this inference benchmark doesn’t support MCUs and other resource-limited platforms due to the absence of small benchmarks and compatible implementations. To meet this need, \cite{tiny-mlperf} introduced MLPerf Tiny, the first industry-standard benchmark suite for ultra-low-power tiny machine learning systems. However, the benchmark only considers 4 simple tasks including Keyword Spotting, Visual Wake Words, Image Classification on CIFAR10 and Anomaly Detection. These are basic models and as of now they don’t support more complicated tasks such as object detection.

EEMBC's CoreMark \cite{CoreMark} has gained popularity as the standard benchmark for MCU-class devices, thanks to its user-friendly implementation and application of real-world algorithms. However, it doesn’t fully profile entire programs or accurately depict machine learning inference workloads. The consortium introduced the MLMark benchmark\cite{MLMark} to fill this gap, but the models tested are still too demanding for MCU devices. In a similar vein, YOLO models have previously been benchmarked for embedded systems with YOLOBench\cite{yolobench}. This work tests 550+ YOLO-based object detectors on four datasets and hardware platforms but this benchmark is applicable for edge devices with higher footprints as compared to tiny MCUs.  

Several other works \cite{stereo_labs,opencv_yolo_benchmarking,feng2022benchmark,zhu2022performance} have proposed methods to benchmark the performance of various architectures from the YOLO series on server-grade and embedded GPUs, as well as specialized accelerator target platforms. Again, we observe that models from these benchmarks are not suitable for MCU devices. 

We have seen that existing benchmark studies either offer models for extremely simple deep learning tasks, or models which exceed the memory and compute constraints of MCUs. There is an evident and clear need for a TinyML benchmark that accommodates a range of complex models on complex datasets, enabling a more equitable comparison and benchmarking of MCUs for more cutting edge industrial use cases.

\begin{figure*}[ht!]
    \centering
    \includegraphics[width=0.99\textwidth]{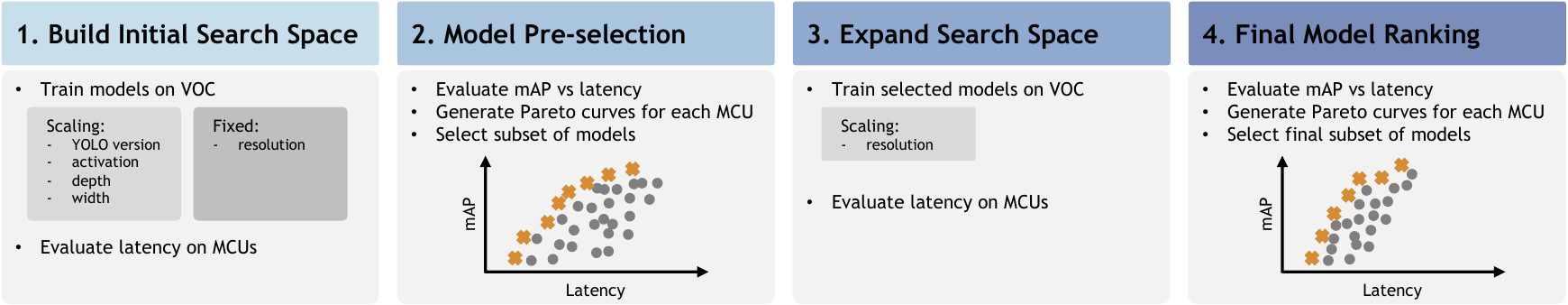}
    \caption{Flowchart of the \textit{MCUBench} process for model candidate generation, pre-selection and ranking. Pareto-optimal points are depicted as orange crosses. }
    \label{fig:scheme}
\end{figure*}

\section{Methodology}
\label{sec:method}

\begin{table}[ht]
\centering
\caption{\textit{MCUBench} architecture space (variation of backbone/neck, depth, width, activation function and input resolution).}
\label{tab:mcubench_space}
\vspace*{2mm}
\begin{tabularx}{.9\linewidth}{X|X|X|p{4.8cm}}
\toprule
\textbf{Model} & \textbf{Backbone} & \textbf{Neck} & \textbf{Depth Factors}\\
\midrule
YOLOv3 \cite{redmon2018yolov3} & DarkNet53 & FPN & \{0.125, 0.16, 0.20, 0.25\} \\
YOLOv4 \cite{bochkovskiy2020yolov4} & CSPDNet53 & SPP-PAN & \{0.125, 0.16, 0.20, 0.25\}\\
YOLOv5 \cite{Jocher_YOLOv5_by_Ultralytics_2020} & CSPDNet53-C3 & SPPF-PAN-C3 & \{0.125, 0.16, 0.20, 0.25\}\\
YOLOv6s-3 \cite{li2023yolov6} & EfficientRep & RepBiFPAN & \{0.085, 0.125, 0.16, 0.20, 0.25\}\\
YOLOv7 \cite{wang2023yolov7} & E-ELAN & SPPF-ELAN-PAN & \{\}\\
YOLOv8 \cite{ultralytics} & CSPDNet53-C2f & SPPF-PAN-C2f & \{0.20, 0.25\} \\
\bottomrule
\end{tabularx}
\vspace*{1mm}
\begin{tabularx}{\linewidth}{X}
\centering
Activation Function $\in$ \{ReLU, SiLU\} \\
Input resolution $\in$ \{128, 160, 192, 224\} \\
Width factor $\in$ \{0.05, 0.085, 0.125, 0.16, 0.20, 0.25\} \\
\end{tabularx}
\end{table}

YOLO detectors can be differentiated based on their backbone, neck, and head architectures. Other important factors, such as the total number of parameters, training pipeline, hyperparameters, choice of loss, and activation function, also contribute to improving the performance of these models. This benchmark studies the effect of the backbone, neck, activation function, and total parameters (width, block depth, and input resolution) on evaluation metrics like average precision, latency, and RAM requirements. To highlight the influence of the selected factors, the detection head architecture, training pipeline, loss function, and hyperparameters are kept fixed throughout the benchmark process.

For the current benchmark process on MCU devices, we use a decoupled detection head from YOLOv8~\cite{yolo_survey}. This head is chosen for its anchor-free nature, which provides latency benefits in end-to-end detection pipelines~\cite{rtdetr}. The same loss function as YOLOv8 (CIoU and DFL losses) is used to predict bounding boxes. The training pipeline and hyperparameters are borrowed from the Ultralytics~\cite{ultralytics} code due to its simplicity and ability to reproduce state-of-the-art (SOTA) results.

For designing the candidate architectures of the YOLO detectors, the current work varies the constituent backbone and neck from the available options while keeping the head fixed from YOLOv8. Usually, the scaling variants considered for these architectures are from ($n$, $s$, $m$, and $l$) but even the smallest variation $YOLOv\{family\}n$ is more than 2 MB in size (INT8 quantized model) and it does not fit in MCU memory constraints which necessitates the use of smaller scaling variants. These scaling factors and choice of activation function can influence the latency, RAM, Flash and mAP of these models.
While the Sigmoid-Weighted Linear Units (SiLU)\cite{SiLU} activation function is generally used for detection tasks across all YOLO families to achieve state-of-the-art (SOTA) results, we utilized both ReLU~\cite{ReLU} and SiLU activation functions to address both average precision and latency aspects. The flow of candidate model generation, along with possible variations, is shown in~\ref{fig:scheme}.

We utilized six mainstream YOLO family models (YOLOv3 to YOLOv8) with six variations in width factors and two variations in activation function. Each YOLO family has different variations in block depth factors due to differences in their maximum block depth. For example, in YOLOv7, the depth of a block is always fixed at $1$, so multiplying any depth scaling factors will not produce new variants. Additionally, we used four variations of input resolution from $128\times128$ to $224\times224$, with a step of $32$, to generate the final Pareto frontier models.
For YOLOv6, v3.0 provides different architectures for $s$, $m$, and $l$ variations. Although the current work proposes custom variations in width and depth scaling factors and activation function variations, these could have been introduced on all the original $s$, $m$, and $l$ architectures. However, for this benchmark process, the YOLOv6 v3.0 $s$ architecture is used as the base form, over which other variations in input resolutions, width/depth factors, and activation functions are introduced.

\subsection{Hardware Benchmarking}

\begin{table*}[ht!]
\centering
\caption {Specifications of MCUs used for benchmarking. {\footnotesize *Note: The total memory may be higher for some boards; only the memory allocated for AI applications is listed.}}
\label{tab:transposed_benchmarking_hardware}
\resizebox{\linewidth}{!}{
\begin{tabularx}{\linewidth}{p{5.0cm}|p{4.0cm}|>{\centering\arraybackslash}p{1.2cm}>{\centering\arraybackslash}p{1.2cm}|>{\centering\arraybackslash}p{1.2cm}>{\centering\arraybackslash}p{1.2cm}}
\toprule
\textbf{Device} & \textbf{ARM Cortex Core} & \multicolumn{2}{c|}{\textbf{Flash} [kB]} & \multicolumn{2}{c}{\textbf{RAM} [kB]} \\
& & \textbf{Int} & \textbf{Ext} & \textbf{Int} & \textbf{Ext} \\
\midrule
NUCLEO-H743ZI\cite{stm32_nucleo} & M7 @480 MHz & 2048  & N/A & 1024  & N/A \\
B-U585I-IOT02A\cite{stm32_iot_kit} & M33 @160 MHz & 2048  & 65536 & 768* & N/A \\
STM32F469I-DISCO\cite{stm32_f469NI} & M4 @180 MHz & 2048  & 16384 & 384  & 16384 \\
STM32F769I-DISCO\cite{stm32_f769NI} & M7 @216 MHz & 2048  & 65536 & 512  & 16384 \\
STM32H573I-DK\cite{stm32_h573} & M33 @250 MHz & 2048  & 65536 & 640  & N/A \\
STM32H747I-DISCO\cite{sth747i_disco_board} & M4+M7 @400 MHz & 1024* & 131072 & 704* & 8192 \\
STM32L4R9I-DISCO\cite{stm32_l4r9} & M4 @120 MHz & 2048  & 65536 & 640  & N/A \\
\bottomrule
\end{tabularx}}
\end{table*}

The actual inference latency for each model can vary significantly depending on the RAM, Flash, processor type, and clock frequency. Therefore, we collected the memory and latency measurements for each model by running inferences on seven different boards with MCUs, as described in Table~\ref{tab:transposed_benchmarking_hardware}.

\subsubsection{Hardware Details}
We selected a diverse range of boards with MCUs to capture various performance characteristics. The STM32H7 series, including the STM32H747I-DISCO~\cite{sth747i_disco_board} and NUCLEO-H743ZI~\cite{stm32_nucleo}, are included for their high-performance capabilities. The STM32F4 and STM32F7 series, represented by the STM32F469I-DISCO~\cite{stm32_f469NI} and STM32F769I-DISCO~\cite{stm32_f769NI}, are known for their advanced graphics and display support. The B-U585I-IOT02A~\cite{stm32_iot_kit} and STM32H573I-DK~\cite{stm32_h573}, featuring Cortex-M33 processors, offer secure connectivity and are suitable for various applications. Finally, the STM32L4 series, like the STM32L4R9I-DISCO~\cite{stm32_l4r9}, is selected for its ultra-low power consumption. Each MCU offers different configurations of RAM, Flash, processor type, and clock frequency, allowing us to evaluate their performance across a variety of models.

\subsubsection{Benchmarking Approach}
We used the ST Microelectronics developer cloud~\cite{st_dev_cloud} to gather memory and latency information by running the models on actual hardware. The ST Developer cloud exposes a RestAPI, which we utilized to automate the collection of memory and latency data for multiple models across multiple MCUs in parallel. To compile the models from TFLite, we used ST Microelectronics' latest XCubeAI tool~\cite{st_xcubeai} (version 9.0). XCubeAI is a cloud hosted tool which allows selection of different MCUs, compile and test AI models without any need to have the actual physical device. This virtualization allowed us to run hundreds of models on a range of different MCUs. This is the main reason of using all the MCUs from ST family (with different speed and memory constraints). Models initially in the ONNX format \cite{onnx} (FP32) were converted and quantized to TFLite \cite{tflite} (INT8) using the~\texttt{onnx2tf}~\cite{onnx2tf} library. During this conversion, we ensured that both input and output were kept in UINT8 format and applied per-tensor quantization. This approach was chosen to optimize performance while maintaining compatibility with the hardware. All latency measurements were performed with a batch size of $1$, averaged over multiple inference cycles. We measured the inference time required to execute the YOLO model graph, including bounding box decoding post-processing operations performed after the last convolutional layers of the network. Other bounding box post-processing steps, such as non-maximum suppression, were excluded from these measurements.

\subsubsection{Memory Consumption Analysis: Flash and RAM}
Internal Flash memory is used to store the model's weights and architecture. In cases where the model size exceeds the available internal Flash, execution will fail if the MCU relies only on internal memory. However, MCUs with external memory can load larger models without failure. For instance, models that could not fit in the internal Flash of the NUCLEO-H743ZI were successfully executed on the STM32H747I-DISCO due to its external memory support. RAM is utilized for storing intermediate computations and activations during inference. Higher RAM capacity enables more efficient handling of these computations, reducing latency. MCUs with external RAM support can handle higher resolution inputs and complex models without memory limitations. For example, we noticed that models with higher resolution inputs failed on the B-U585I-IOT02A but ran successfully on the STM32H747I-DISCO, which has more RAM (internal and external combined) and demonstrated better performance metrics compared to MCUs with less RAM.

\subsubsection{Impact on Latency}
The amount of available RAM directly impacts the inference time of the models. MCUs with higher RAM typically show reduced latency due to the ability to store intermediate computation results more efficiently. For instance, the STM32H747I-DISCO, with its higher RAM capacity, demonstrated significantly lower latency compared to the STM32L4R9I-DISCO, which has comparatively less RAM. This reduction in latency is crucial for real-time applications where prompt responses are necessary. Flash memory variations also affect latency, particularly in terms of loading and initializing the model weights and architecture. MCUs with larger Flash memory, such as the NUCLEO-H743ZI, exhibited faster model initialization and lower overall latency. This is due to the reduced need for frequent memory accesses and the ability to store more of the model's parameters internally, minimizing the delays associated with external memory accesses. Clock cycle variations play a significant role in determining the inference time of the models. Higher clock frequencies generally lead to faster processing speeds, as more instructions can be executed per second. For example, the NUCLEO-H743ZI, operating at a higher clock frequency, showed improved latency performance compared to the STM32H747I-DISCO, which operates at a slightly lower clock frequency. The increased clock speed allows for quicker data processing and shorter execution times, which is particularly beneficial for computationally intensive tasks within the YOLO model.

\subsection{Model Selection Procedure}
We now present the four main steps of our benchmarking procedure, all summarized in Figure~\ref{fig:scheme}. First, we (1) built the initial search space by fixing the resolution and scaling between YOLO family, activation, depth, and width. Then, we (2) preselected Pareto models for each hardware to select a subset. Next, we (3) scaled the resolution, and finally, we (4) generated the final Pareto models.

To evaluate potential model candidate architectures, we used the PASCAL VOC dataset~\cite{pascal-voc-2012}, which includes 20 object categories. All 240 models were trained from scratch using the Ultralytics YOLOv8~\cite{ultralytics} training pipeline, default VOC hyperparameters, and DFL \& CIoU losses for 100 epochs. We trained these models at a higher resolution (448×448) to use them as pretrained models for fine-tuning at smaller resolutions. No pretrained weights, such as backbone pretraining on ImageNet, were used.

\subsubsection{Step 1 -- Build Initial Search Space} 
The first step involves generating the initial set of models by training $240$ models at a resolution of $448\times448$ and then fine-tuning them at a resolution of $192\times192$ for $10$ epochs. At this stage, we kept the resolution fixed at $192\times192$ while scaling the YOLO version, activation function, depth, and width. We then evaluated the models by exporting them to TFLite and running them on seven different MCUs to gather RAM, Flash, and latency data. It is important to note that not all models were able to run on all MCUs due to memory constraints, unsupported operations, etc. The actual number of models with available benchmark results is mentioned in Table~\ref{tab:mcubench_space}. For example, the NUCLEO-H743ZI was able to run only $96$ out of $240$ models in this step. All models with benchmark results and mAP values constitute the search space models.

\subsubsection{Step 2 -- Model Pre-selection}
In the second step, models were preselected based on their latency vs mAP Pareto optimality. This was done for each device to build the seven Pareto curves. Only the Pareto-optimal solutions for each MCU were selected to proceed to further steps $3$ and $4$.
For this procedure, the mAP@$50$ values from all variants fine-tuned at a resolution of $192\times192$ were used, along with the latency measurements obtained from performing inference on all available MCU platforms. We identified Pareto-frontier models for each device based on the list of models that ran smoothly on that platform. Finally, we merged the lists from all devices to form the "First Pareto Set." This set contains a total of $159$ models across different devices, of which $72$ were unique across all the boards (e.g., $21$ Pareto models were selected from the NUCLEO-H743ZI MCU, as shown in Table\ref{tab:mcubench_space}).

\subsubsection{Step 3 -- Expand Search Space}
At this stage, the models selected from Steps 1 and 2 were used to generate an expanded search space. We scaled the input resolution to three more levels apart from $192 \times 192$: $128 \times 128$, $160 \times 160$, and $224 \times 224$ and further fine-tuned these models on the VOC dataset. This fine-tuning happened for $10$ epochs, starting from the trained weights of $448 \times 448$. Mosaic-augmentation was disabled after the 5\textsuperscript{th} epoch to follow the Ultralytics training pipeline. While more epochs could enhance model convergence, $10$ epochs were deemed sufficient for benchmarking.

\subsubsection{Step 4 -- Final Model Ranking} 
In the final stage, the models from Step 3 were evaluated on the seven MCUs to build new Pareto curves. We ran 72 models at four different resolutions ($128$, $160$, $192$ and $224$) on all seven devices and collected benchmark results.

\section{Results}
~\label{sec:results}
The benchmarking procedure consisted of four main steps: building the initial search space, model pre-selection, expanding the search space, and generating the final Pareto curves. We started with the evaluation of $240$ YOLO model variants on 7 different MCUs, generated Pareto frontiers of $72$ models, expanded the search space to $288$ models by training at different resolutions, and finally analyzed the $131$ Pareto-optimal models. Below, we present the results for each step in detail.

\begin{table}[h]
    \centering
    \caption{Summary of results for Pareto-optimal models across various MCU platforms. The table presents the number of models at each stage of the benchmarking process.}
    \label{tab:step_bystep_summary}
    \begin{tabularx}{0.8\linewidth}{>{\raggedright\arraybackslash}p{4.5cm}|>{\centering\arraybackslash}X|>{\centering\arraybackslash}X|>{\centering\arraybackslash}X|>{\centering\arraybackslash}X}
    \toprule
    \textbf{Device} & \textbf{Step 1} & \textbf{Step 2} & \textbf{Step 3} & \textbf{Step 4} \\
    \midrule
    NUCLEO-H743ZI & 96 & 21 & 156 & 41 \\
    B-U585I-IOT02A & 226 & 23 & 254 & 27 \\
    STM32F469I-DISCO & 36 & 12 & 71 & 49 \\
    STM32F769I-DISCO & 85 & 23 & 165 & 47 \\
    STM32H573I-DK & 117 & 29 & 181 & 47 \\
    STM32H747I-DISCO & 149 & 28 & 164 & 50 \\
    STM32L4R9I-DISCO & 109 & 23 & 208 & 34 \\
    \midrule
    Total Models & 818 & 159 & 1191 & 296 \\
    Unique Pareto Models & & \textbf{72} & & \textbf{131} \\
    \bottomrule
    \end{tabularx}
\end{table}

For clarity, we provide the number of models for each of the steps mentioned above, as shown in Table~\ref{tab:step_bystep_summary}. Notably, the number of models in Step 1 is less than 240 for each device, as some models failed to run, because of the presence of not-yet supported operators in the architecture by the board, or exceeding memory (Flash and RAM) limitations. For eg. even after NUCLEO-H743ZI has highest clock frequency among all the devices, we could only run 96 models because of the absence of external Flash. Models beyond 2048 kB size failed to run on this device. B-U585I-IOT02A has most of the models supported as it has external Flash to fit more models.  

\subsection{Model Pre-Selection}
\begin{figure*}[ht!]
    \centering
    \includegraphics[width=1\textwidth]{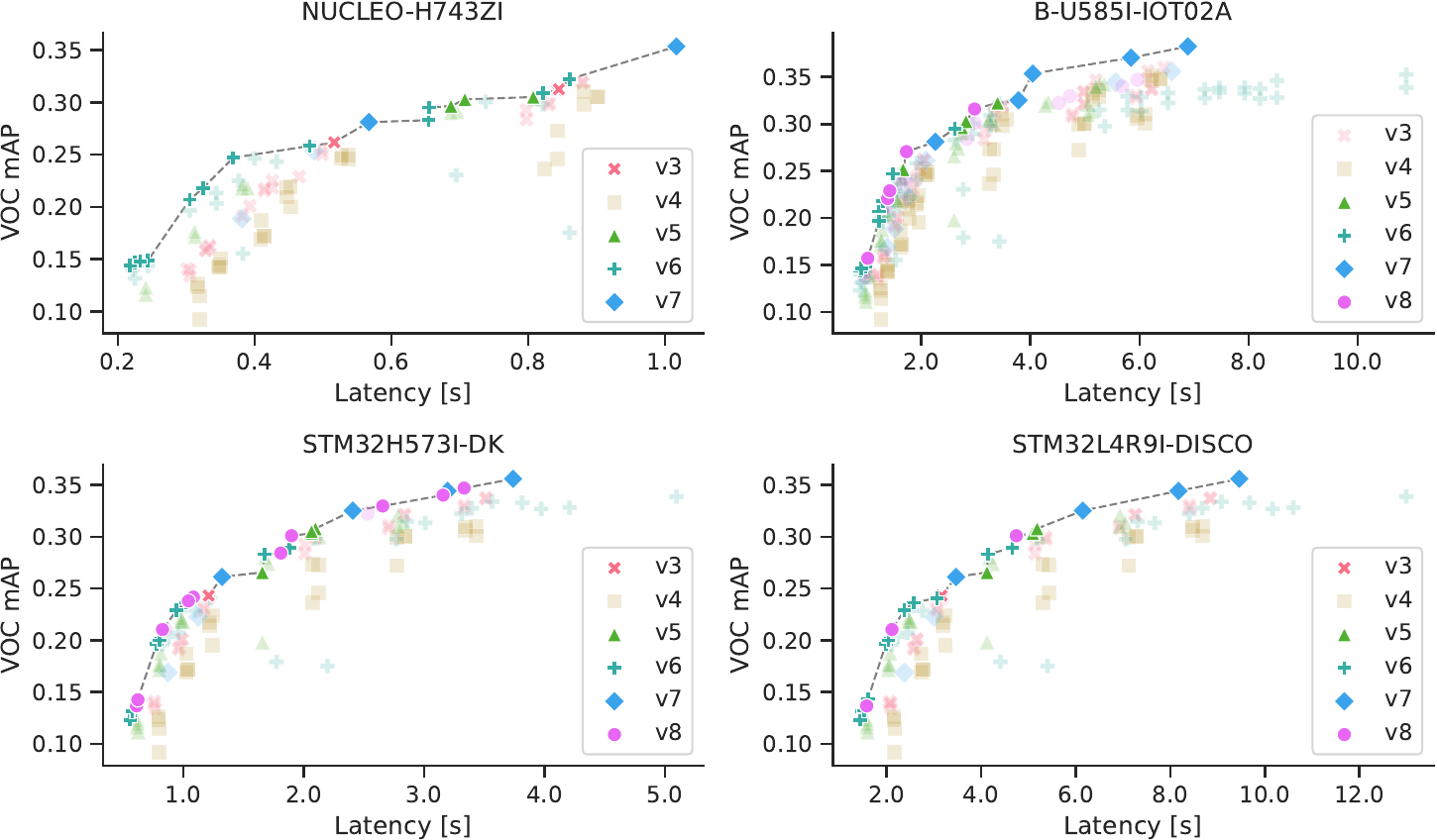} %
    \vspace{2mm}
    \caption{Pareto frontiers of \textit{MCUBench} models trained on the VOC dataset (on several target resolutions) 4 different hardware platforms. Each point represents a single model in the mAP-latency space, with the model family coded with color and marker shape (all YOLOv6-3.0 models are represented by the same color).}
    \label{fig:step1_4mcus}
\end{figure*}
The initial step involved evaluating $240$ YOLO model variants on various MCUs. The Pareto frontiers were generated by comparing mean Average Precision (mAP) versus latency for each model, resulting in a subset of models for each device. Figure~\ref{fig:step1_4mcus} shows the Pareto fronts for 4 MCUs, while the detailed results for all 7 MCUs are reported in the Appendix (Figure~\ref{fig:step1_all_mcus}). Different YOLO model families are represented by distinct marker types and colors, with the Pareto frontier highlighted by dashed lines. We observe that the types of Pareto-optimal models vary by device. For example, the NUCLEO-H743Z has a large number of v6 models, while the B-U585I-IOT02A and STM32H573I-D show a majority of v6, v7, and v8 models in the Pareto curve.

It is also evident that the absolute latency values vary significantly across devices. For instance, the maximum latency observed for the NUCLEO-H743ZI is around 1 second, whereas the STM32L4R9I-DISCO experiences latencies up to 12 seconds. This disparity highlights the different performance capabilities and constraints of the various MCUs.

In Step 1, the total number of models tested across all devices was 818. However, as shown in Table~\ref{tab:step_bystep_summary}, the Pareto frontiers were computed for each device to filter the models down to 159. After combining the results from all devices to identify the unique Pareto-optimal models, we found 72 unique models. These models were then used for the subsequent steps (Step 3 and Step 4) mentioned in section $3.2$.

\subsection{Pareto Analysis: mAP vs Latency}
In this section, we present the results from steps 3 and 4 (section $3.2$) of our benchmark flow. Initially, we selected 72 unique models and fine-tuned them for additional epochs on the VOC dataset at four different resolutions. We then evaluated the latency of these models on 7 different hardware platforms.

The total number of models evaluated in step 4 was 1191. From these, we selected the Pareto-optimal models for each device, resulting in 296 models. However, only 131 of these models were unique across all devices, as shown in Table~\ref{tab:step_bystep_summary}.

\begin{figure*}[ht!]
    \centering
    \includegraphics[width=0.99\textwidth]{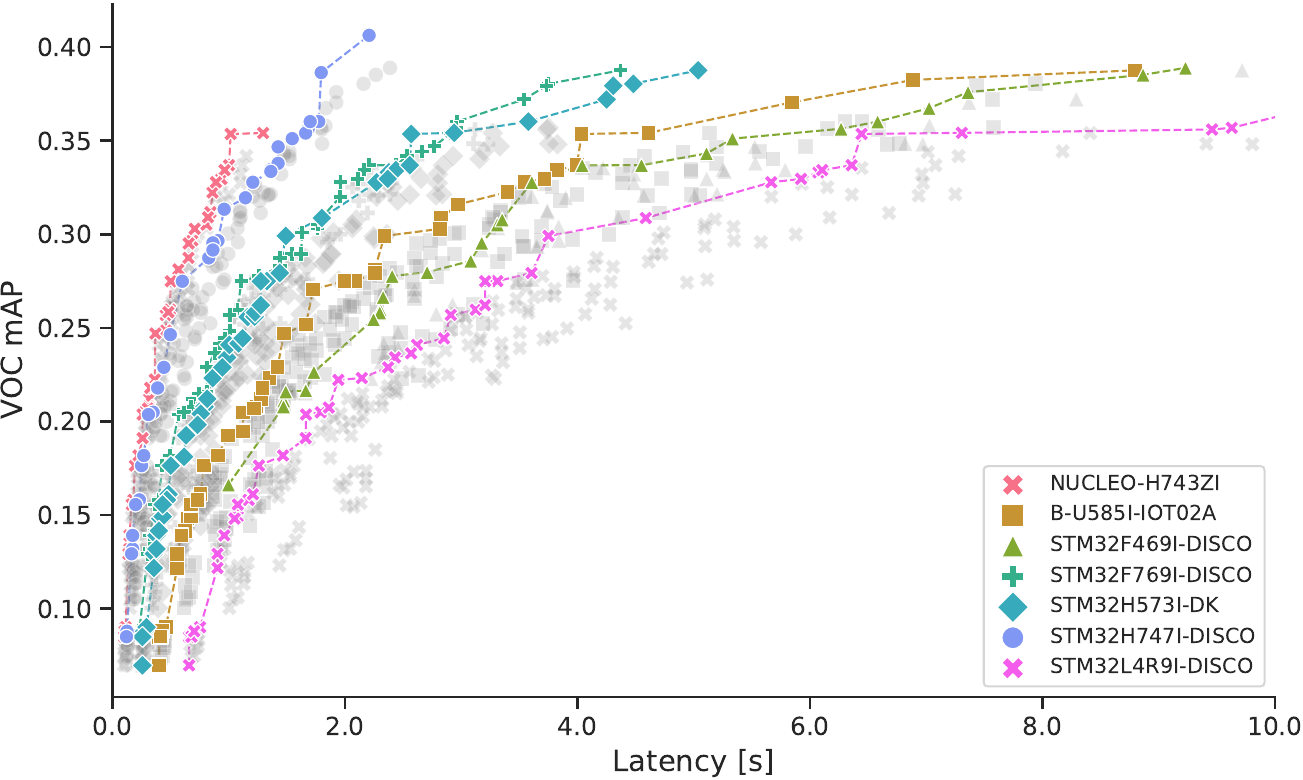} %
    \caption{Combined Pareto Fronts for All Devices. This plot illustrates the Pareto-optimal models on the VOC dataset across various MCU hardware platforms. The x-axis represents latency (in seconds), and the y-axis represents mAP. Each marker corresponds to a different hardware platform, with the faded markers representing all tested models and the solid markers representing the Pareto-optimal models. The dashed lines connect the Pareto-optimal models for each device. The details of the minimum and maximum latency solutions for each device are summarized in Table~\ref{tab:pareto_frontiers}.}
    \label{fig:step4_parerto}
\end{figure*}

In Figure~\ref{fig:step4_parerto}, we present the Pareto frontiers for all 7 devices. Each marker corresponds to a different device, as indicated in the legend. This visualization allows us to compare the performance and efficiency of the models across various hardware platforms. The Pareto curves highlight the trade-off between mean Average Precision (mAP) and latency, with the goal of identifying models that achieve high accuracy with minimal latency. The plot shows that NUCLEO-H743ZI is the most powerful compared to others because the latency of the most accurate model is around one second and STM32L4R9I is the least powerful as the latency for highest mAP model is almost 10 seconds. STM32F769I-DISCO and STM32F769I-DISCO and comparable devices as Pareto curve for these two have a strong correlation. Also, apart from dependence on model complexity, latency also depends greatly on the extent of use of external memory. For eg. B-U585I-IOT02A with 764 KB RAM might perform better for larger models compared to STM32F769I-DISCO with 512 KB RAM even if the former is less powerful (160 MHz vs 216 MHz). 

Table~\ref{tab:results_model_family_hw} shows the distribution of YOLO model families for each board evaluated. Interestingly, even older YOLO architecture families can produce optimal models when trained using modern techniques. For example, the B-U585I-IOT02A and STM32L4R9I-DISCO boards include 15 and 10 YOLOv3 models, respectively, in their Pareto lists, out of a total of 27 and 34 Pareto models. Conversely, only 1 and 3 YOLOv8 models appear in the Pareto lists for same two devices. YOLOv5 models are evenly placed in Pareto list of all devices and YOLOv4 is not present in any of these devices. 

\begin{table}[h]
\centering
\caption{Distribution of Pareto models: number of models per family and activation function across different devices.}
\label{tab:results_model_family_hw}
\vspace*{2mm}
\begin{tabular}{>{\raggedright\arraybackslash}p{3.5cm}|>{\centering\arraybackslash}p{.7cm}>{\centering\arraybackslash}p{.7cm}>{\centering\arraybackslash}p{.7cm}>{\centering\arraybackslash}p{.7cm}>{\centering\arraybackslash}p{.7cm}>{\centering\arraybackslash}p{.7cm}|>{\centering\arraybackslash}p{.9cm}>{\centering\arraybackslash}p{.9cm}|>{\centering\arraybackslash}p{1cm}}
\toprule
\textbf{Device} & \multicolumn{6}{c|}{\textbf{Model Family}} & \multicolumn{2}{c|}{\textbf{Activation}} & \multicolumn{1}{c}{\textbf{Total}}  \\
\textbf{} & {v3} & {v4} & {v5} & {v6} & {v7} & {v8} & {ReLU} & {SiLU}  & {}\\
\midrule
NUCLEO-H743ZI & 2  & 0  & 8  & 24  & 4  & 3 & 6 & 35 & \textbf{41}\\ 
B-U585I-IOT02A & 15 & 0  & 9  & 2   & 0  & 1 & 10 & 17 & \textbf{27}\\ 
STM32F469I-DISCO & 0  & 0  & 9  & 17  & 6  & 17 & 13 & 36 & \textbf{49}\\ 
STM32F769I-DISCO & 0  & 0  & 8  & 23  & 6  & 10 & 12 & 35 & \textbf{47}\\ 
STM32H573I-DK & 0  & 0  & 10 & 19  & 6  & 12 & 5 & 42 & \textbf{47}\\ 
STM32H747I-DISCO & 0  & 0  & 10 & 25  & 5  & 10 & 25 & 25 & \textbf{50}\\ 
STM32L4R9I-DISCO & 10 & 0  & 1  & 16  & 4  & 3 & 4 & 30 & \textbf{34}\\ 
\bottomrule
\end{tabular}
\vspace*{1mm}
\end{table}

The YOLOv6 family is the most common in the Pareto lists, predominantly due to its ability to produce small, fast models that achieve minimal latency. This is more a reflection of the model configuration, specifically the deeper maximum block size that allows for effective depth scaling, rather than an inherent superiority of the YOLOv6 architecture. Interestingly, the B-U585I-IOT02A board include just 2 of YOLOv6 models in its Pareto list, highlighting the variability in performance across different hardware configurations. This also suggests that the benefits seen in YOLOv6 models on other boards are not universally applicable and depend heavily on the specific hardware characteristics. Figure \ref{fig:step1_4mcus} also shows that most of the lower latency models are each hardware is from YOLOv6 family due to the reason mentioned above. 

Table~\ref{tab:pareto_frontiers} shows the minimum and maximum latency solutions for each device. YOLOv6 models are predominant in all minimum latency solutions, while maximum latency solutions include most of the models YOLOv7 family. There is one exception for STM32F469I-DISCO where both minimum and maximum latency solution is from YOLOv3. This indicates that while YOLOv6 models excel in achieving low latency, they do not necessarily outperform other models in all aspects. The resolution analysis follows the general rule that lower resolutions result in faster models and higher resolutions in slower models.The fastest model in our benchmark (with 0.08 mAP) has 0.10 latency on NUCLEO-H743ZI device and the most accurate model has 0.41 mAP with latency of 2.21 seconds on STM32H747I-DISCO (on VOC dataset).

\begin{table}[h]
\centering
\caption{Summary of Pareto frontiers for each device. This table presents the Pareto-optimal \textit{MCUBench} models on the VOC dataset across various MCU hardware platforms. For each device, the models are selected based on achieving the best mAP$_{50-95}$ under given latency constraints (both minimum and maximum latency). The scaling parameters (depth and width factors) are denoted as dXXwYY, where dXX means depth factor $=0.XX$ and wYY means width factor $=0.YY$. The input resolution of the models is also reported.}
\label{tab:pareto_frontiers}
\vspace*{2mm}
\begin{tabularx}{\linewidth}{l|cccc|cccc}
\toprule
\textbf{Device} & \multicolumn{4}{c|}{\textbf{Min. Latency Solution}} & \multicolumn{4}{c}{\textbf{Max. Latency Solution}} \\
 & \textbf{Model} & \textbf{Scaling} & \textbf{Res.} & \textbf{(Lat., mAP)} & \textbf{Model} & \textbf{Scaling} & \textbf{Res.} & \textbf{(Lat., mAP)} \\
\midrule
NUCLEO-H743ZI & v6 & d85w50 & 128 & (0.10, 0.08) & v7 & d250w160 & 224 & (1.30, 0.35) \\ 
B-U585I-IOT02A & v6 & d85w50 & 128 & (0.40, 0.07) & v7 & d250w250 & 224 & (8.80, 0.39) \\ 
STM32F469I-DISCO & v3 & d160w125 & 128 & (1.00, 0.17) & v3 & d250w250 & 224 & (9.23, 0.39) \\ 
STM32F769I-DISCO & v6 & d85w50 & 128 & (0.22, 0.08) & v7 & d250w250 & 224 & (4.37, 0.39) \\ 
STM32H573I-DK & v6 & d85w50 & 128 & (0.26, 0.07) & v7 & d250w250 & 224 & (5.04, 0.39) \\ 
STM32H747I-DISCO & v6 & d85w50 & 128 & (0.12, 0.08) & v7 & d250w200 & 224 & (2.21, 0.41) \\ 
STM32L4R9I-DISCO & v6 & d85w50 & 128 & (0.66, 0.07) & v7 & d250w250 & 224 & (12.80, 0.39) \\ 
\bottomrule
\end{tabularx}
\vspace*{1mm}
\end{table}
\subsection{Scaling Parameters Analysis: Depth, Width, Resolution, Activation}
\begin{figure*}[ht!]
    \centering
    \includegraphics[width=1\textwidth]{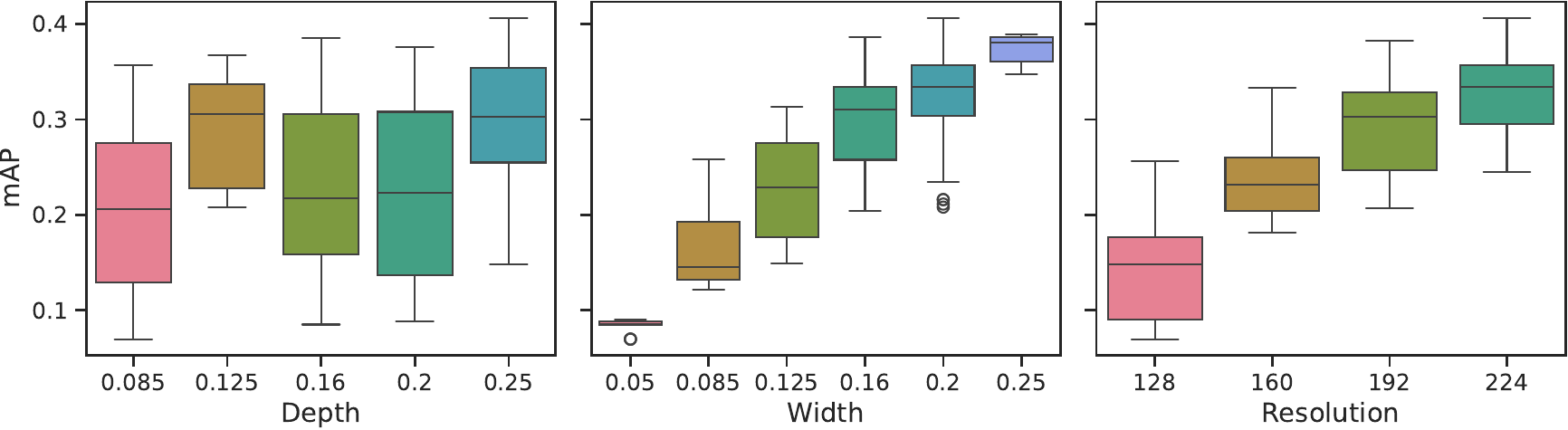}
    \caption{Statistics of model scaling parameters (depth factor, width factor, input resolutions) in Pareto-optimal models on VOC (Step 4) across 7 different MCUs. The plot highlights trends for width and resolution, where increased values correspond to higher mAP, while depth shows less consistent improvement.}
    \label{fig:pareto_scaling}
\end{figure*}

To further establish the best parameters to effectively scale when deploying object detectors on MCUs, we analyzed the impact of three main parameters in the final Pareto models. The boxplots in Figure~\ref{fig:pareto_scaling} illustrate the impact of depth, width, and resolution on mAP.

It is clear from the boxplots that increasing resolution and width generally leads to higher mAP values, indicating their strong influence on model performance. For instance, increasing the resolution from 128 to 224 results in more than double the mAP. Similarly, increasing the width factor from 0.05 to 0.25 shows a significant improvement in mAP. This consistency in improvement highlights the importance of optimizing resolution and width for enhancing model performance.

In contrast, the depth parameter shows more variability and a less consistent impact on mAP. The mAP values for depth factors ranging from 0.085 to 0.25 do not exhibit a clear trend of improvement. This suggests that depth adjustments are less effective compared to resolution and width optimizations. The variability in depth impact could be due to the increased complexity and potential overfitting, which might not translate to better performance on constrained devices like MCUs.

Figure~\ref{fig:pareto_scaling} shows that the mAP increases from approximately 0.1 at a resolution of 128 to over 0.35 at a resolution of 224. Similarly, increasing the width factor from 0.05 to 0.25 results in a substantial increase in mAP, reaching above 0.4 for the highest width factor. Depth, on the other hand, does not show a significant upward trend, with mAP values fluctuating between 0.1 and 0.3 across different depth factors.

Changing Activations does not affect FLASH and MACs but it affect latency moderately (SiLU is slower compared to ReLU) but RAM SiLU model consumes considerably more RAM and it can adversely affect latency when internal RAM is not sufficient.

Overall, these observations suggest that focusing on optimizing resolution and width is more beneficial for improving model performance on MCUs than adjusting depth. This insight is crucial for developing efficient object detection models that perform well under the constraints of MCU hardware.

\section{Conclusion}
In this work, we present $MCUBench$, a latency-mAP benchmark of hundreds of YOLO-based models specifically designed for constrained devices like MCUs (less than 4 MB quantized models). We apply our method to 7 MCUs with different specifications, using models trained on the VOC object detection dataset. The mAP, latency, RAM, and Flash data are collected in a fixed, controlled environment with independent variables of model width, block depth, activation function, and input image resolution. The data suggests interesting characteristics of these models on different MCUs. Our model scaling results reveal that optimizing resolution and width is more effective for enhancing model performance compared to block depth adjustments. $MCUBench$ is the first benchmark to combine the mainstream task of object detection with the strict constraints of MCU deployment. This benchmark can aid in selecting models for a target device and accuracy budget without the need for costly model training and evaluation. It can also compare the performance of different MCUs on the same set of models.

\section{Future Work}
Currently, $MCUBench$ models are trained on the VOC dataset to create a large search space for model selection. However, VOC data is simpler compared to mainstream datasets like COCO~\cite{lin2014microsoft}. Future work will train Pareto models on the COCO dataset for more generalized models and explore simpler object detection tasks, such as person detection, which are suitable for cheaper MCUs in smart home devices. This study currently uses only STMicroelectronics boards, but future work could include boards from other vendors like Sony, Renesas and Infineon. We also plan to add models generated by compression methods and compare them with those generated using the search space defined in this paper.

\section*{Acknowledgements}

We would like to extend our sincere gratitude to Ivan Lazarevich and Saptarshi Mitra, whose contributions during their time with us were instrumental in the early stages of this research. These initial ideas were critical in developing MCUBench. \\
We would like to thank ST Microelectronics.  XCubeAI, their publicly available developer cloud allowed us to automate this extensive benchmarking.


\newpage
\section*{Appendix}

\begin{figure*}[ht!]
    \centering
    \includegraphics[width=0.85\textwidth]{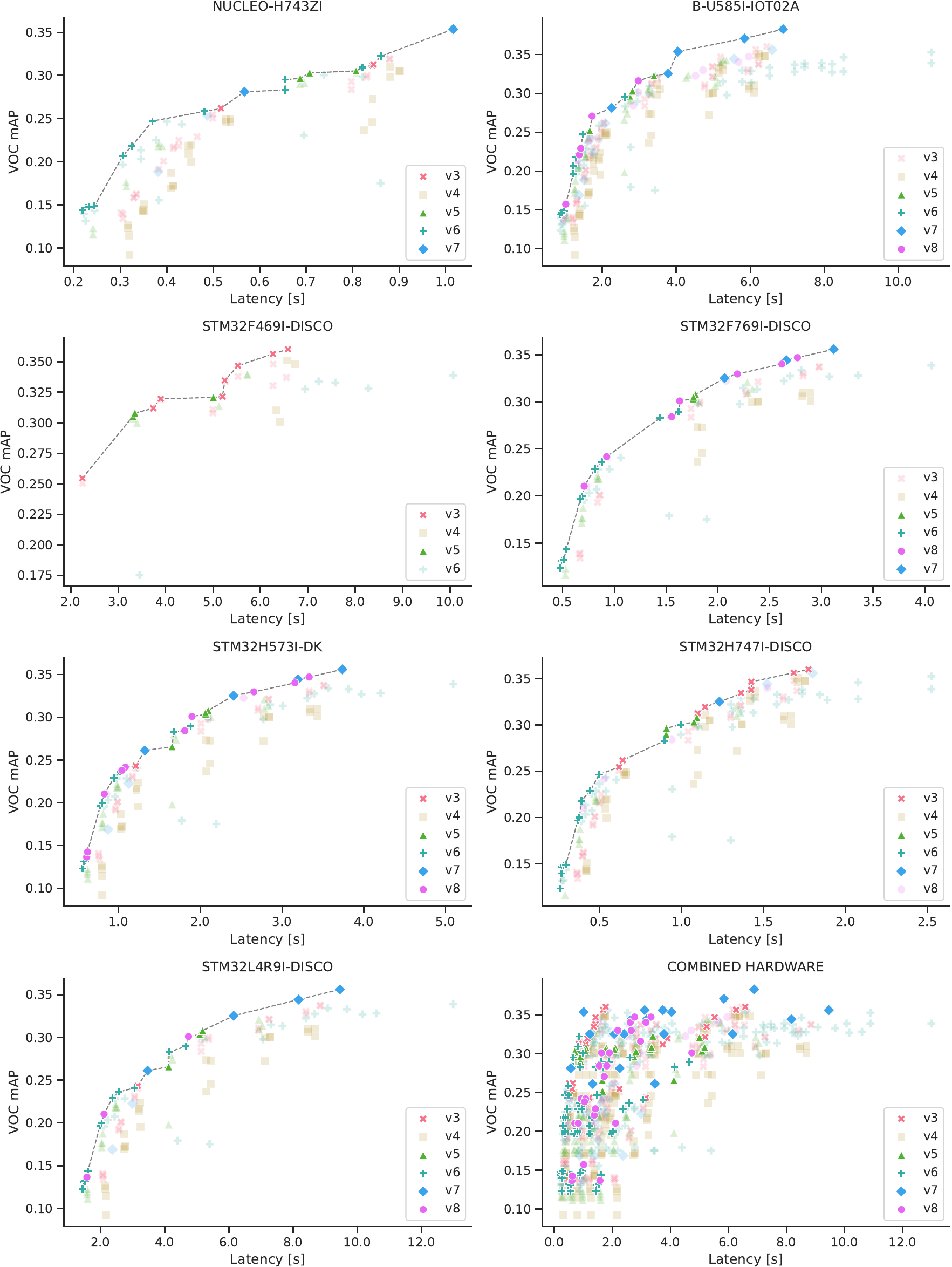} %
    \caption{Combined Pareto frontiers of \textit{MCUBench} models fine-tuned on the VOC dataset at several target resolutions on 7 different MCUs. Each point represents a single model in the mAP-latency space, with the model family coded with color and marker shape.}
    \label{fig:step1_all_mcus}
\end{figure*}

In this appendix, we provide supplementary details to support the main findings of our study, including a comprehensive figure illustrating the combined Pareto fronts for all 7 MCUs in the pre-selection step (Figure~\ref{fig:step1_all_mcus}).

\end{document}